# The Absent-Minded Driver Problem Redux

Subhash Kak

*Abstract*

This paper reconsiders the problem of the absent-minded driver who must choose between alternatives with different payoff with imperfect recall and varying degrees of knowledge of the system. The classical absent-minded driver problem represents the case with limited information and it has bearing on the general area of communication and learning, social choice, mechanism design, auctions, theories of knowledge, belief, and rational agency. Within the framework of extensive games, this problem has applications to many artificial intelligence scenarios. It is obvious that the performance of the agent improves as information available increases. It is shown that a non-uniform assignment strategy for successive choices does better than a fixed probability strategy. We consider both classical and quantum approaches to the problem. We argue that the superior performance of quantum decisions with access to entanglement cannot be fairly compared to a classical algorithm. If the cognitive systems of agents are taken to have access to quantum resources, or have a quantum mechanical basis, then that can be leveraged into superior performance.

INTRODUCTION

The problem of rational choice with imperfect recall was introduced by Kuhn [1] and the matter of the absent-minded driver was examined by Piccione and Rubenstein [2] as a single person game. The driver must have a strategy to return home choosing from different exit ramps that have varying payoff. The absent-minded driver has no memory and this represents a special case of the Bayesian problem of choosing between actions with state-dependent payoff [3].

Since the decisions taken at different intersections do not depend on the previous one (due to the absent-mindedness of the driver) this problem belongs to the class of decision problems with incomplete information and it has bearing on the general area of rational agency, learning, social choice, mechanism design, auctions theories of knowledge, and belief. It is also of significance in such extensions of classical decision theory that include quantum processes.

Piccione and Rubenstein (PR) thought that the problem was associated with a paradox for upon arrival at an intersection the driver will be able to see the situation somewhat differently than what he had planned before he began driving. This assumption is problematic for what absentmindedness means is not clearly defined and all that the driver must be allowed is probabilistic decision at each intersection. Aumann *et al* [3] pointed out that the claim of the paradox was from an incorrect comparison of the planning stage to the action stage. They rightly concluded that while the considerations at the planning and action stages do differ, there was no paradox or inconsistency.

But what if the driver had less information at the planning stage than allowed by PR? From an information theoretic perspective this situation should lead to a payoff that is lower than obtained by PR for their example. What if the driver is absentminded but he has a consistent strategy that is hardwired into his decision system? We show that with such an assumption the driver will do better than if he had more initial information. This may be viewed as a trade-off between two different kinds of information and such problems belong to the general area of decision-making in the presence of limited information and uncertainty [4].

If one were to consider repetitive plays of the game with respect to nodes that are listed in rank order, the behavior of the driver may be seen to parallel social network formation where connections to existing nodes do not depend on previous connections. The generation of language where the speaker or the writer picks letters may also be viewed as similar to the exits on the highway. We find power laws characterize a whole range of such phenomena [5]. In the general case, the various exits may be labeled 1, 2, … n, in a graph characterized by some value function whose definition is driven by extraneous considerations of theory.

It has been suggested that in a general decision problem a quantum decision machine may be used instead of a classical algorithm. This is done most simply by making available to the driver accessory quantum resources [6]. In general, quantum models of cognition and information processing [7]-[13] appear to provide new insight into the workings of human agents in different decision environments (e.g. [10][13]). According to the orthodox interpretation of quantum theory, cognitive capacity cannot be explained using any ontology based on physical structures [12]. The consideration of quantum cognition opens up the possibility that the quantum resource of entanglement underlies paradoxical behavior of human agents in social networks and, in turn, it could be exploited to provide superior performance in engineered and business systems.

If the absent-minded driver problem is compared to that of finding a specific item within a random list (due to the absent-mindedness), it is known that a quantum algorithm will always be superior to the best classical algorithm. With *n* items in the database, the Grover algorithm solves this problem in $O(\sqrt{n})$ steps as against $O(n)$ steps needed for the best classical algorithm [14][15]. One would expect, therefore, that a quantum decision system will always be superior although it should be realized that the instrumentation of the quantum system comes with its own problems [16][17][18].

Returning to the absent-minded driver, it is clear that if each intersection were associated with a qubit and the probability of exit is related to the probability of finding, say a 0, this is identical to the use of a classical probabilistic algorithm. When the qubits are entangled then the performance of the quantum decision device is known to be superior [6]. But this entails a resource that is way more than what is available to the driver who only uses classical computation.

Although it was argued in [6] that the use of a quantum decision tree improves the payoff, the kind of resources compared for the classical and the quantum cases are different. If it is assumed that the car (if not the driver) keeps track of the number of the exit, the classical algorithm will be as good.

In classical decision theory the actors are classical agents. Now that scholars are considering quantum models of cognition and decision, it is reasonable to reexamine the assumptions about the nature of the processing by the agents. At the other end, one can keep to the classical model of the rational agent but provide the driver with ability to make quantum computations.

This paper reexamines the problem of the absent-minded driver. It is shown that a particular non-uniform assignment strategy for successive choices does better than a fixed probability strategy and this is true even if the payoffs of each are unknown. Both classical and quantum approaches to the problem are considered. It is argued that the resources needed for quantum decisions cannot be fairly compared to those in a classical algorithm. Access to quantum resources results in superior performance.

THE ORIGINAL PROBLEM

We revisit the problem of the absent-minded driver as originally visualized [2] which we call Example 1:

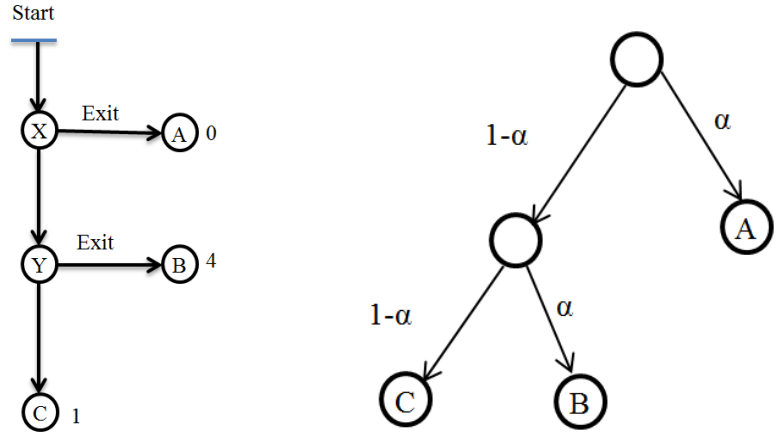

Figure 1. The path of the driver and the corresponding graph (Example 1)

A driver is sitting in a bar planning his trip home. In order to get there he must take the highway where the first exit (A; payoff 0) is hazardous and to be avoided. The second exit (B; payoff 4) is home. If he misses both exits he'll reach the end of the highway and have to spend the night at a hotel (C; payoff 1). If he exits always at A, his payoff is 0 and if he exits always at B, it is 4. Because of absent-mindedness, his payoff will be within these two limits.

Since the driver does not know which intersection he is at, it was assumed by PR [2] that he should use a consistent strategy where the probability of his exiting is α and that of not exiting is 1- α. The payoff for the driver is therefore $P_{1U}$ (where 1 in the subscript refers to 1 of Example 1):

$$P_{1U} = 4(1-\alpha)\alpha + (1-\alpha)^2 \tag{1}$$
$$= 1 + 2\alpha - 3\alpha^2$$

$\dfrac{dP_{1U}}{d\alpha} = 2 - 6\alpha = 0$ which yields that the optimum value of α=1/3.

Thus

$P_{1U}$= 4/3 (2)

The answer to the question as to what additional resources are needed to get a payoff higher than 4/3 would depend on two factors:

- The information on the individual node payoffs in not clearly known
- The amount of memory available to the driver is different from what is taken above

We now modify the problem slightly. We add another node D in Figure 2 that has a payoff of 1.

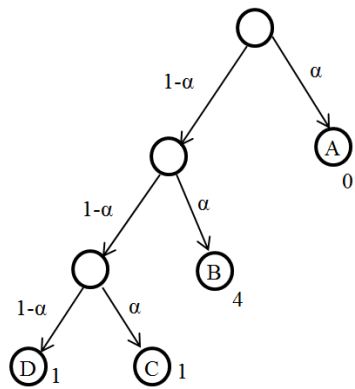

Figure 2. A larger example with fixed probabilities (Example 2)

Again assuming that the driver is absentminded (has zero memory) and he uses a probabilistic strategy that is computed at the planning stage, the payoff is:

$$P_{2U} = 4(1-\alpha)\alpha + (1-\alpha)^2 \alpha + (1-\alpha)^3$$
$$= 1 + 2\alpha - 3\alpha^2$$

$\dfrac{dP_{2U}}{d\alpha} = 2 - 6\alpha = 0$ implies that, as before, α=1/3.

$P_{2U}$= 4/3 (3)

The payoff for this strategy is also 4/3. The way Example 1 was enlarged into Example 2, did not change the payoff and the probabilistic decision related to exiting from the highway.

# UNKNOWN PAYOFFS WITH ONE ITEM OF MEMORY

Now consider unknown payoffs. What should the driver do? Let the unknown payoffs of destinations A, B, C, and D be a, b, c, and d. It will be assumed that the driver knows how many exits exist and let their number be *k* (which is 4 for Example 2). This number was known at the planning stage in the previous case as well.

We would expect that any consistently applied decision strategy will do worse than the previous case. If not, it would mean that the lack of knowledge of the individual payoffs has been compensated by the assumed additional capacity of keeping track of the number of exits.

From a Bayesian perspective, the best performance will be obtained for the case of unknown payoffs if each of them is reached with the same probability.

Assume that the driver adopts the strategy of reaching each of the exits with the same probability. This is equivalent to the following algorithm:

---

**Decision Algorithm with Unknown Payoffs and *k* Exits**

1. Take the first exit with probability $\frac{1}{k}$
2. Take the second exit with probability $\frac{1}{k-1}$
3. ---
4. The last two exits will have probabilities of $\frac{1}{2}$

---

The probabilities that will be associated with the exits on the fly will be as shown in Figure 3.

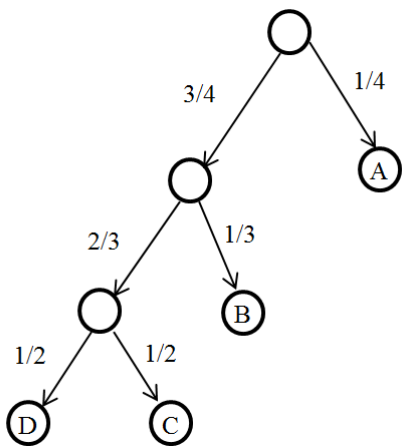

Figure 3. The branching probabilities for Example 2.

The total payoff is

$$P_{2NU} = (a+b+c+d)/4 = 6/4 = 3/2 \qquad (4)$$

We observe that

$$P_{2NU} > P_{2U} \qquad (5)$$

The application of this algorithm to Example 1 yields a payoff of 5/3, which is better than 4/3 by the random but consistent procedure provided in [2].

## GENERALIZATION TO *k* CHOICES OUT OF *n*

We now consider a generalization of the absent-minded driver problem to that of the student who must register for, say, 2 courses out of a given set of 4 (these numbers can be arbitrary as *k* and *n*, so long as *k* < *n*).

The problem is, therefore, of going through the graph of Figure 2 twice over while ensuring that the first choice is excluded in the second run. Let the payoff represent the value of the course as published in the students' social network.

Using the fixed probability method, one needs to develop the optimum payoff for the next round by taking out the node that was selected first.

For the second round, the true payoff will be obtained by averaging over all the cases. Let the payoff in the second round with choice of k in the first round be P(k). Then the total payoff is

$$P_{\text{Total}} = \sum_i \frac{1}{n}(P(i) + \sum_{j=0}^{n-1} P(j|i)) \qquad (6)$$

Thus for Example 2 where the payoffs for the selection problem for the 4 cases are 0, 4, 1, 1, the choices are as follows:

*First choice A.*
Payoff 0. The second choice payoff is 4α+(1-α)α+(1-α)(1-α)=1+3α.

   Total payoff = 1+3α

*First choice B.*
Payoff 4. The second choice payoff is 0+ (1-α)α+(1-α)(1-α)=1-α

   Total payoff = 5-α

*First choice C.*
Payoff 1. The second choice payoff is 0+ 4(1-α)α+(1-α)(1-α)

Total payoff = 2+2α -3α²

*First choice D.*
Payoff 1. The second choice payoff is 0+ 4(1-α)α+(1-α)(1-α)

$$\text{Total payoff} = 2+2\alpha -3\alpha^2 \tag{7}$$

The total average is = ¼ (10+6α -6α²)

The optimum value is for -12 α+ 6=0 or α=1/2.

$$\text{The optimum is } \frac{1}{4}(10+6/2-6/4) = 23/8 \tag{8}$$

If one were to use non-uniform allocation, the corresponding values that will be obtained are:

0+2, 4+2/3, 1+5/3, 1+5/3

$$\text{Their average is } \frac{1}{4}(8+4) = 3 \tag{9}$$

The improvement in performance by using non-uniform probability as against uniform probability is

$$3-23/8 = 1/8 \tag{10}$$

## QUANTUM DECISION NETWORK

Assume that the intersections are indistinguishable to the driver and he has access to a qubit he can test at each intersection. Then depending on if he gets a 0 on his measurement he exits otherwise he continues on the highway. Let the qubit be described by the state [6]:

$$|\phi\rangle = \sqrt{\alpha}|0\rangle + \sqrt{(1-\alpha)}|1\rangle \tag{11}$$

The absent-minded driver will then exit consistently with the probability of α and the case is identical to that of Figure 2.

Now let us assume that in Example 1 the resource available to the driver includes entangled qubits. At the planning stage the driver has the map of the intersections and he can order the kind of qubits he wants. Specifically, let us assume that he chooses qubits at the first two intersections that are fully entangled as below:

$$|\phi\rangle = \frac{1}{\sqrt{2}}(|01\rangle + |10\rangle) \tag{12}$$

Then there is a 50% chance that the first qubit is 0, in which case he will exit with a payoff of 0. But if the first qubit is 1 (stay on the highway), then the second qubit that is measured at the following intersection will definitely be 0 (exit) with the corresponding payoff of 4. The average payoff would thus have increased from the 4/3 of the random algorithm of PR [2] to 2 for the entangled qubits method.

Suppose the high payoff exit was the third one rather than the second and it was assumed that the driver was befuddled about the exact intersection. The following entangled state will ensure that if he did not exit on the first intersection, he will not exit on the second one as well, but will definitely exit on the third.

$$|\phi\rangle = \frac{1}{\sqrt{2}} (|001\rangle + |110\rangle) \tag{13}$$

The payoff will be the average of the payoffs of the first and third exits.

Alternatively, qubits may be designed in this fashion:

$$|\phi\rangle = |110\rangle \tag{14}$$

This will ensure that the driver stays on the highway for the first two exits and takes the third exit. But this assumes that the qubits code the information of the exit number in the state function. A similar counting of the exits in the classical case also will guarantee the optimum result.

## DISCUSSION

This paper reconsidered the problem of the absent-minded driver who must choose between alternatives with different payoff with imperfect recall and varying degrees of knowledge of the system. The classical version of the problem represents the case with minimal information and it has bearing on the general area of rational agency, learning, social choice, mechanism design, auctions, and theories of knowledge [19][20]. Within the framework of extensive games, this has applications to many artificial intelligence scenarios.

It is obvious that the performance of the agent improves as the memory available increases. It was shown that a non-uniform assignment strategy for successive choices does better than a fixed probability strategy and the result for unknown payoffs with non-uniform assignment does better than uniform assignment with known payoffs.

We considered both classical and quantum approaches to the problem. We argued that the superior performance of quantum decisions with access to entanglement cannot be fairly compared to a classical algorithm. If the cognitive systems of agents are taken to have access to quantum resources, then that can be leveraged into superior performance.